\documentclass[letterpaper,10pt,conference]{ieeeconf}

\IEEEoverridecommandlockouts
\usepackage{graphicx}
\usepackage{amsmath}
\usepackage{amssymb}
\usepackage{booktabs}
\usepackage{multirow}
\usepackage{cite}
\usepackage{url}

\usepackage[T1]{fontenc}
\usepackage{graphicx} % Required for inserting images
\usepackage{booktabs}
\usepackage[table, dvipsnames]{xcolor}
\usepackage{amssymb}
\usepackage{amsmath}
\usepackage{soul}
\usepackage{color}   %May be necessary if you want to color links
\usepackage{hyperref}
\hypersetup{
    colorlinks=true, %set true if you want colored links
    linktoc=all,     %set to all if you want both sections and subsections linked
    linkcolor=blue,  %choose some color if you want links to stand out
}
\usepackage{multirow}

\usepackage{tabularx}
\usepackage{xltabular}
\usepackage{ragged2e}
\usepackage{array}

\usepackage{tikz}
\usetikzlibrary{mindmap, backgrounds}
\usepackage[font=footnotesize]{caption}
\usepackage{stfloats}
\title{\LARGE \bf
Does Continual Imitation Learning Remain Grounded?\\
A Language-Perturbed Benchmark for Robotic Task Retention}

\author{
Siddeshwar Raghavan$^{1}$, Ziqin Yuan$^{2}$, Fengqing Zhu$^{1}$, and Byung-Cheol Min$^{3}$%
\thanks{$^{1}$Department of Electrical and Computer Engineering, Purdue University, West Lafayette, IN, USA.
{\tt\small \{raghav12,zhu0\}@purdue.edu}}%
\thanks{$^{2}$Department of Computer and Information Technology,Purdue University, West Lafayette, IN, USA.
{\tt\small yuan460@purdue.edu}}%
\thanks{$^{3}$Computer Science and Intelligent Systems Engineering, Indiana University Bloomington, IN, USA.
{\tt\small minb@iu.edu}}%
}

\begin{document}
% ============================================================

\maketitle

% ------------------------------------------------------------
% Abstract
% ------------------------------------------------------------
\begin{abstract}
Continual imitation learning evaluates whether a robot can learn new knowledge without forgetting previously learned skills. However, retaining task performance does not ensure the behavior remains grounded in language because policies may rely on scene cues, object associations, or memorized task structure. We introduce a benchmark protocol to study how language-guided behavior changes as robotic policies learn successive tasks. We construct meaning-preserving and meaning-changing instruction variants for the Goal, Spatial, Object, and Long suites of LIBERO. Policy experiments focus on LIBERO-Goal, evaluating Original and Paraphrase instructions after each continual-learning stage. We compare representative continual imitation learning methods under their original assumptions while separating task competence from language sensitivity. The proposed diagnostics complement standard learning and forgetting metrics by measuring semantic robustness, goal adaptation, and language sensitivity. Results show that strong continual-learning performance does not always translate to reliable language grounding, and our diagnostics help determine whether retained skills remain correctly guided by their instructions. Additional materials are available at \url{https://sites.google.com/view/stillgrounded}.
\end{abstract}

%%%%%%%%%%%%%%%%%%%%%%%%%%%%%%%%%%%%%%%%%%%%%%%%%%%%%%%%%%%%%%%%%%%%%%%%
%%%%%%%%%% INTRO %%%%%%%%%%%%%%%%%%%%%%%%%%%%%%%%%%%%%%%%%%%%%%%%%%%%%%%
%%%%%%%%%%%%%%%%%%%%%%%%%%%%%%%%%%%%%%%%%%%%%%%%%%%%%%%%%%%%%%%%%%%%%%%%

\section{Introduction}

Robot policies conditioned on language must acquire new behaviors while retaining previously learned skills. Continual imitation learning studies this challenge through streams of incoming manipulation tasks. Benchmarks such as LIBERO~\cite{liu2023libero} characterize acquisition and retention using task success, forward transfer, negative backward transfer, area under the continual performance curve, and final success rate. However, these metrics do not establish whether retained behavior remains appropriately guided by language.

\begin{figure}[t]
    \centering
    \includegraphics[width=\linewidth]{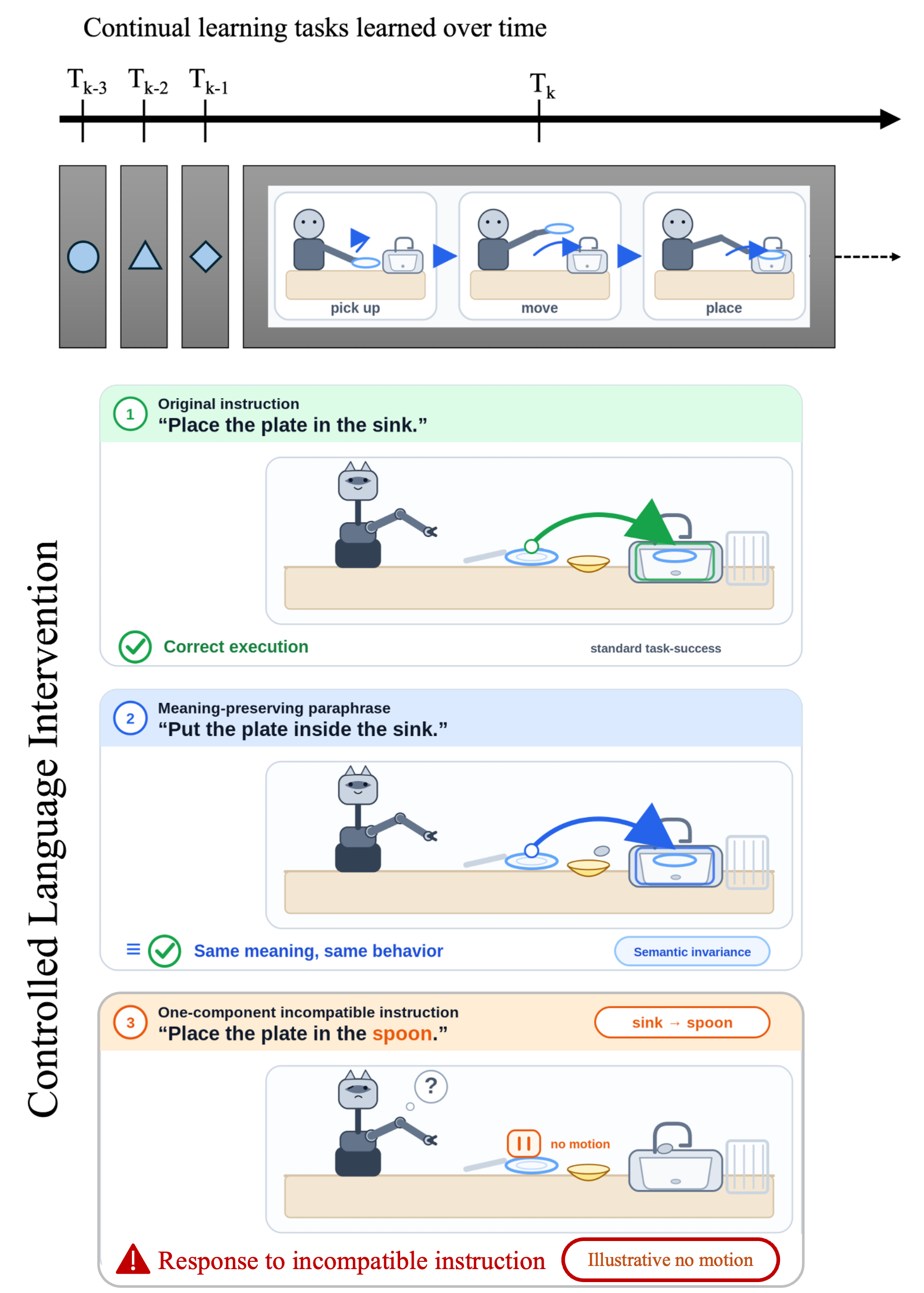}
    \vspace{-10pt}
    \caption{\textbf{Task retention does not guarantee language grounding.} Original instructions and paraphrases should lead to the same behavior, while semantic changes may require a different response. The final panel shows one possible response to an incompatible request. Our protocol evaluates these behaviors across continual learning checkpoints rather than relying only on original task success.}
    \label{fig:intro}
    \vspace{-20pt}
\end{figure}

This distinction matters because a policy may select behaviors using scene layout, object associations, or memorized task routing rather than the instruction itself. Figure~\ref{fig:intro} illustrates why task retention alone does not establish language grounding. Success under a paraphrase does not necessarily indicate language use, since the policy may behave identically while ignoring the instruction. Likewise, a behavioral change does not guarantee completion of the modified goal, which may also be infeasible in the current scene. We therefore test whether equivalent instructions preserve behavior, valid goal changes induce the appropriate response, and policies respond appropriately when no valid goal can be executed.

Recent robustness benchmarks such as LIBERO PRO~\cite{zhou2025liberopro} and LIBERO-Plus~\cite{liberoplus} show that strong task performance does not always mean that a policy is using language reliably. A robot may still succeed by relying on scene layout, object associations, or other shortcuts. In continual learning, this raises an additional question: as new tasks are learned, does the connection between language and behavior remain stable?

We study this by evaluating saved continual learning checkpoints with fixed changes to the instruction. We consider four cases, the \textbf{original instruction}, a \textbf{paraphrase} with the same meaning, a \textbf{minimal semantic change} that requests a different goal, and a collision that makes the request incompatible with the current scene. These cases let us test whether a policy preserves behavior when meaning stays the same, changes behavior when the requested goal changes, and responds sensibly when no valid goal can be executed. Not every instruction change can be evaluated in the same way. Paraphrases can be compared against the original task goal, while valid minimal contrasts can be evaluated against the modified goal. For incompatible requests, there may be no correct success target, so we instead examine behavioral responses such as whether the policy continues toward a known goal, initiates motion, or changes its action distribution. This avoids treating every instruction change as a simple success-or-failure test. 

Our analysis is performed after training using policies saved at successive continual-learning stages. This allows us to study how language robustness changes over time without modifying the training process itself. We also report whether each method uses task identity or explicit routing, since strong retention through task-specific modules is different from behavior selected through language alone.

Although our experiments use LIBERO dataset, the protocol decouples instruction construction and evaluation from benchmark-specific scene and goal definitions, making it possible to extend the same analysis to other continual robot-learning settings and benchmarks.

We summarize our contributions below:
\begin{enumerate}
    \item We introduce a temporal evaluation protocol for studying how language guided behavior changes across successive continual learning checkpoints.
    \item We use goal-based metrics when an instruction defines a valid goal, and behavioral diagnostics when it does not.
    \item We establish protocol coverage across all four LIBERO suites and evaluate policies on LIBERO-Goal, comparing representative continual imitation learning methods without modifying their algorithmic design.
\end{enumerate}

%%%%%%%%%%%%%%%%%%%%%%%%%%%%%%%%%%%%%%%%%%%%%%%%%%%%%%%%%%%%%%%%%%%%%%%%
%%%%%%%%%% REL WORK %%%%%%%%%%%%%%%%%%%%%%%%%%%%%%%%%%%%%%%%%%%%%%%%%%%%
%%%%%%%%%%%%%%%%%%%%%%%%%%%%%%%%%%%%%%%%%%%%%%%%%%%%%%%%%%%%%%%%%%%%%%%%

\vspace{-3pt}
\section{Related Work}
\vspace{-3pt}

\subsection{Continual robotic imitation learning}

Continual imitation learning studies the learning of successive tasks from demonstrations while mitigating catastrophic forgetting and interference of previously learned knowledge~\cite{khetarpal2020continual_rl,CL_RL_def, woczyk2021continualworldroboticbenchmark}. LIBERO~\cite{liu2023libero} provides a common evaluation benchmark for this problem through data suites that vary goals, objects, spatial relationships, and task complexity.

Continual Learning algorithms are broadly separated into replay~\cite{raghavan2024ocil, ER, raghavan2024delta},
regularization~\cite{Kirkpatrick_2017, SI, MAS}, parameter isolation~\cite{rusu2022progressiveneuralnetworks,pathnet, expertgate}, and architectural expansion \cite{khetarpal2020continual_rl,lopezpaz2017gem}.
Sequential fine tuning provides a baseline without an explicit mechanism for preventing forgetting. Elastic Weight Consolidation (EWC)~\cite{Kirkpatrick_2017} penalizes changes to parameters estimated to be important for previous tasks, while PackNet~\cite{mallya2018packnetaddingmultipletasks} allocates parameters through iterative pruning and protects previously assigned weights. These general continual learning strategies provide useful reference points for robotic imitation.

However, continual learning has evolved to use pretrained models for a better set of representations as the starting point~\cite{wang2022dualprompt, wang2022learning, lamdapilot, codaprompt, raghavan2026hearlocalizesegmentcontinually}. In the case of continual imitation learning, TAIL~\cite{liu2024tail} investigates adapters, L2M~\cite{schmied2023learning} was introduced for reinforcement learning and uses a learned modulation pool to adapt a frozen pretrained model. IsCiL~\cite{lee2024incremental} develops retrievable skills for continual adaptation, while DMPEL~\cite{dmpel} progressively constructs a library of
parameter efficient experts and dynamically combines them through routing. We evaluate implementations of these approaches within a common robotic imitation framework. Rather than proposing another mechanism for retaining skills, our work examines whether the behaviors retained by these different mechanisms remain appropriately dependent on language. 

\subsection{Language robustness and diagnostic evaluation}

Language-conditioned imitation policies use natural-language instructions to guide robot manipulation across different objects and tasks~\cite{shridhar2022cliport,jang2022bcz}. Robustness benchmarks evaluate behaviors that standard task success may miss. LIBERO-PRO~\cite{zhou2025liberopro} studies changes in objects, initial states, instructions, and environments, while LIBERO-Plus~\cite{liberoplus} extends this evaluation to broader visual, physical, and language variations. Contrast Sets~\cite{contrast_sets} use small, targeted changes to reveal how sensitive language-guided policies are to changes in the evaluation setting.

%%%%%%%%%%%%%%%%%%%%%%%%%%%%%%%%%%%%%%%%%%%%%%%%%%%%%%%%%%%%%%%%%%%%%%%%
%%%%%%%%%% METHODS  %%%%%%%%%%%%%%%%%%%%%%%%%%%%%%%%%%%%%%%%%%%%%%%%%%%%
%%%%%%%%%%%%%%%%%%%%%%%%%%%%%%%%%%%%%%%%%%%%%%%%%%%%%%%%%%%%%%%%%%%%%%%%
\vspace{-3pt}
\section{Benchmark Protocol}
\vspace{-3pt}
\label{sec:benchmark_protocol}

\begin{figure*}[t]
    \centering
    \includegraphics[width=0.78\textwidth]{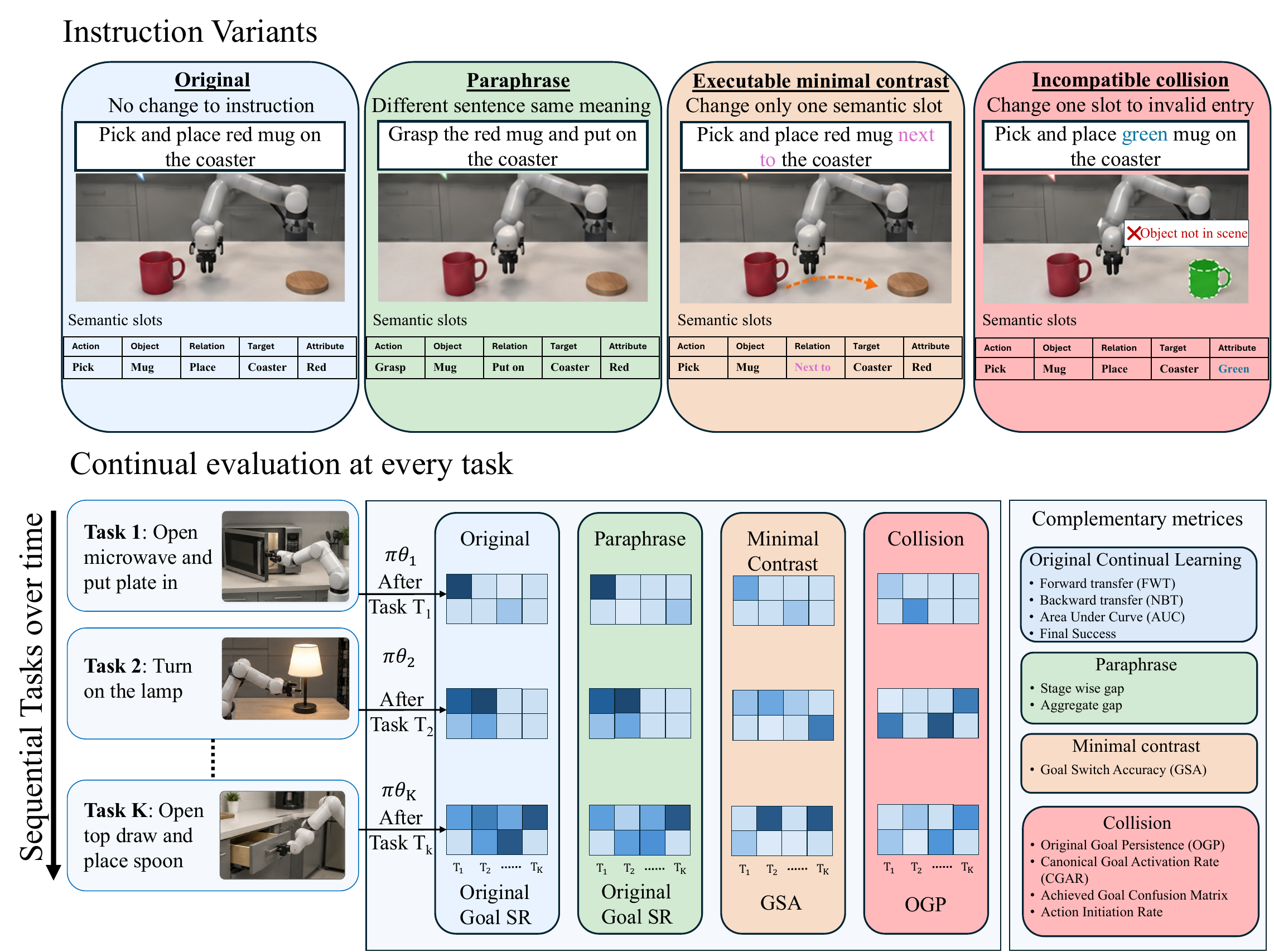}
    \caption{\textbf{Overview of the benchmark protocol:} Original instructions and meaning-preserving paraphrases are evaluated across continual-learning stages, while one-slot minimal contrasts and scene-incompatible collisions assess final-stage goal switching and behavioral sensitivity.}
    \vspace{-15pt}
    \label{fig:benchmark}
\end{figure*}

\subsection{Problem Setting}

Figure~\ref{fig:benchmark} summarizes the benchmark conditions and their evaluation stages. We consider a continuous stream of imitation-learning tasks
$\mathcal{T}_{1:K}=\{\mathcal{T}_1,\ldots,\mathcal{T}_K\}$. At stage $t$, the learner receives demonstrations
$D_t=\{(o_n,a_n,l_t)\}_{n=1}^{N_t}$ and updates a language-conditioned policy
\begin{equation}
    a \sim \pi_{\theta_t}(\cdot\mid o,l)
    \label{eq:policy_no_task_id}
\end{equation}
The benchmark does not constrain the retention mechanism. The only replay method, ER, retains past demonstrations. SeqFT retains only its updated parameters, while EWC additionally stores parameter-importance statistics. PackNet preserves task-specific parameter masks and has routing ID. L2M and TAIL retain lightweight modulation modules or adapters. IsCiL maintains retrievable skills, and DMPEL progressively expands an expert library with learned routing.

After learning through stage $t$, we evaluate checkpoint $\pi_{\theta_t}$ on every observed task $j\leq t$ under each instruction condition $c$. Conditions with a well-defined executable goal produce the success tensor
\begin{equation}
    S_{t,j,c}
    =
    \frac{1}{N_{\mathrm{eval}}}
    \sum_{n=1}^{N_{\mathrm{eval}}}
    \mathbb{I}\!\left[g_{j,c}(\tau_n)=1\right],
    \qquad j\leq t,
    \label{eq:success_tensor}
\end{equation}
where $\tau_n$ is a rollout from a paired initial state and $g_{j,c}$ is the goal objective implied by condition $c$. Incompatible instructions have no valid target goal and are therefore represented by behavioral outcome statistics, not
by $S_{t,j,c}$.

\subsection{Dataset-Agnostic Task Interface}

For each task, a dataset adapter exposes
\begin{equation}
    \mathcal{A}(\mathcal{T}_j)
    = \left(l_j,\mathcal{E}_j,\mathcal{P}_j,g_j,\mathcal{V}_j\right),
    \label{eq:dataset_adapter}
\end{equation}
where $l_j$ is the canonical instruction, $\mathcal{E}_j$ is the scene-entity inventory, $\mathcal{P}_j$ is the set of supported goal predicates, $g_j$ is the original task goal, and $\mathcal{V}_j$ is the dataset's rollout-success
evaluator. For LIBERO, these fields are extracted from BDDL task definitions. Other datasets can implement the same interface using their task metadata and success annotations. The adapter is used only by the benchmark construction and evaluator and its metadata is not passed to the policy.

\subsection{Controlled Instruction Construction}

Let the semantic frame of instruction $l_j$ be
\begin{equation}
    z_j=(v_j,o_j,r_j,d_j,q_j),
    \label{eq:semantic_frame}
\end{equation}
where $v_j$, $o_j$, $r_j$, $d_j$, and $q_j$ denote the action, manipulated object, relation, target, and attribute, respectively. We use an LLM to propose natural-language variants and then apply deterministic dataset-side validation. We generated candidate variants using the OpenAI Chat Completions API with the model \texttt{gpt-4.1-mini}. We used temperature $0.7$, top-$p$ $0.9$, a maximum of 2,048 output tokens, and seed 101, requesting five paraphrases and one candidate intervention per semantic slot for every task. The complete system prompt, task-conditioned prompt template, generation configuration, and raw model responses will be included with the benchmark release.

\textbf{Original:} The policy receives $l_j$ and is evaluated against $g_j$

\textbf{Paraphrase:} The policy receives a meaning-preserving rewrite $l_j^{+}$ that changes lexical or syntactic form while preserving every semantic
slot:
\begin{equation}
    z(l_j^{+})=z(l_j)
    \label{eq:paraphrase_constraint}
\end{equation}
The paraphrase is evaluated against the original goal $g_j$. Generated paraphrases are checked for redundancy and undergo an independent semantic audit.

\textbf{Validated minimal contrast:} For slot
$k\in\mathcal{K}=\{v,o,r,d,q\}$, a minimal contrast $l_j^{(k)}$ changes exactly
one semantic component:
\begin{equation}
    z_m(l_j^{(k)})=z_m(l_j)\ \forall m\neq k,
    \qquad
    z_k(l_j^{(k)})\neq z_k(l_j)
    \label{eq:minimal_contrast}
\end{equation}
A minimal contrast is retained only if its required entities occur in
$\mathcal{E}_j$, its modified goal uses predicates supported by
$\mathcal{P}_j$, and the goal annotation can be parsed and registered
by the simulator evaluator. Let $\mathcal{R}_j(g)\in\{0,1\}$ denote
this goal-registration check. We define
\begin{equation}
    \begin{aligned}
        \operatorname{Valid}(l_j^{(k)},\mathcal{T}_j)
        =
        \mathbb{I}\!\big[
        &E(l_j^{(k)})\subseteq\mathcal{E}_j
        \land P(g_j^{(k)})\subseteq\mathcal{P}_j \\
        &\land \mathcal{R}_j(g_j^{(k)})=1
        \big].
    \end{aligned}
\label{eq:contrast_validity}
\end{equation}
These checks establish that the modified goal is scene-compatible, but do not guarantee physical feasibility. Consequently, failure to reach the modified goal may reflect either insufficient instruction grounding or limited task execution capability. If no honest one-slot alternative exists, that task-slot pair is reported as unavailable rather than replaced by an artificial example. For LIBERO-Goal, the retained minimal-contrast set contains 10 action, 8 object, 8 target, and 8 relation variants, for a total of 34. No attribute variant satisfies the one-slot construction and validation criteria, so attribute-specific results are reported as unavailable.

\textbf{Incompatible collision:} A collision $l_j^{\times(k)}$ also changes one slot, but references an absent entity or unsupported predicate. A statically validated collision satisfies

\begin{equation}
    E(l_j^{\times(k)})\nsubseteq\mathcal{E}_j
    \quad\lor\quad
    P(l_j^{\times(k)})\nsubseteq\mathcal{P}_j.
    \label{eq:collision_validity}
\end{equation}
Because no correct alternate rollout exists, collisions are sensitivity probes. Cases whose incompatibility cannot be proven are retained as \emph{unverified behavioral probes} but excluded from the validated-collision aggregate.

\subsection{Continual Grounding Metrics}

We report NBT, AUC, and final success under the Original condition. 

For each seed, $S_{t,j,c}$ averages the $N_{\mathrm{eval}}$
rollouts of each instruction variant and then averages variants
equally within task $j$ and condition $c$. We use task-balanced
aggregation:
\begin{equation}
\begin{aligned}
\mathrm{AUC}_c
&=\frac{1}{K}\sum_{j=1}^{K}\frac{1}{K-j+1}
  \sum_{t=j}^{K}S_{t,j,c},\\
\mathrm{NBT}
&=\frac{1}{K}\sum_{j=1}^{K-1}\frac{1}{K-j}
  \sum_{t=j+1}^{K}
  \left(S_{j,j,\mathrm{orig}}-S_{t,j,\mathrm{orig}}\right),\\
\mathrm{FinalSR}_c
&=\frac{1}{K}\sum_{j=1}^{K}S_{K,j,c},\\
\Delta_{\mathrm{AUC}}^{\mathrm{para}}
&=\mathrm{AUC}_{\mathrm{orig}}
-\mathrm{AUC}_{\mathrm{para}}
\end{aligned}
\label{eq:continual_metrics}
\end{equation}

The stage-wise paraphrase gap is
$\Delta_{t,j}^{\mathrm{para}}
=S_{t,j,\mathrm{orig}}-S_{t,j,\mathrm{para}}$.
All metrics are computed independently per seed and reported as the mean and standard deviation across seeds.

% \begin{equation}
%     \Delta^{\mathrm{para}}_{t,j}
%     =S_{t,j,\mathrm{orig}}-S_{t,j,\mathrm{para}},
%     \qquad
%     \overline{\Delta}^{\mathrm{para}}
%     =\frac{1}{|\Omega|}\sum_{(t,j)\in\Omega}
%       \Delta^{\mathrm{para}}_{t,j},
%     \label{eq:paraphrase_gap}
% \end{equation}
where $\Omega=\{(t,j):1\leq j\leq t\leq K\}$. Values near zero indicate similar success under the two instruction conditions, while negative values indicate higher success under paraphrases. The signed gap should be interpreted alongside absolute success rates.

For executable minimal contrasts, we measure Goal-Switch Accuracy (GSA):
\begin{equation}
    \operatorname{GSA}_{t,j,k}
    =\frac{1}{N_{\mathrm{eval}}}
      \sum_{n=1}^{N_{\mathrm{eval}}}
      \mathbb{I}\!\left[
        g_j^{(k)}(\tau_n)=1\land g_j(\tau_n)=0
      \right]
    \label{eq:gsa}
\end{equation}
GSA requires the policy to achieve the modified goal rather than merely change
its actions.

For an incompatible collision, Original-Goal Persistence (OGP) measures how often the policy still achieves the original task goal:
\begin{equation}
    \operatorname{OGP}_{t,j,k}
    =
    \Pr\!\left[g_j(\tau)=1
      \mid \mathcal{T}_j,l_j^{\times(k)},\pi_{\theta_t}\right]
    \label{eq:ogp}
\end{equation}
The verified comparison set $\mathcal{G}_j$ contains the unique
goal-valid annotations associated with the original instruction and retained minimal contrasts for task $j$. CGAR is the fraction of eligible rollouts that newly satisfy at least one goal in
$\mathcal{G}_j$, excluding goals already satisfied after reset and
simulator warmup:
\begin{equation}
    \operatorname{CGAR}_{t,j,k}
    =
    \Pr\!\left[
      \bigvee_{g\in\mathcal{G}(\mathcal{E}_j)}g(\tau)=1
      \mid l_j^{\times(k)},\pi_{\theta_t}
    \right]
    \label{eq:cgar}
\end{equation}
The achieved-goal distribution assigns each rollout to the original goal, another audited goal, multiple goals, or no recognized goal. Separately, Action-Initiation Rate (AIR) is the fraction of all collision-conditioned rollouts in which either the end-effector moves more than $0.01$\,m from its post-reset position or any gripper joint changes by more than $0.01$. AIR is independent of goal attainment and may therefore overlap with OGP and CGAR.

Finally, we compare the policy outputs under the original and perturbed instructions using two action-level metrics. \textbf{Action Divergence} measures the normalized Euclidean distance between the mean actions predicted for the same observation under the two instructions. We normalize by $2\sqrt{d_a}$, the maximum possible distance for a $d_a$-dimensional action space bounded to $[-1,1]$.

We also measure the expert-action language margin:
\begin{equation} 
    \begin{aligned} 
        \operatorname{LM}_{t,j,k} &=\mathbb{E}_{(o,a)\sim D_j} \Big[ \log\pi_{\theta_t}(a\mid o,l_j) \\ &\qquad -\log\pi_{\theta_t}(a\mid o,l_j^{\times(k)}) \Big] 
    \end{aligned} 
\label{eq:language_margin} 
\end{equation}
A positive margin means that the expert action is more likely under the correct instruction than under the perturbed one. We report these action-level metrics alongside task outcomes, since a change in action alone does not necessarily indicate correct language grounding.

%%%%%%%%%%%%%%%%%%%%%%%%%%%%%%%%%%%%%%%%%%%%%%%%%%%%%%%%%%%%%%%%%%%%%%%%
%%%%%%%%%% EXPERIMENT %%%%%%%%%%%%%%%%%%%%%%%%%%%%%%%%%%%%%%%%%%%%%%%%%%
%%%%%%%%%%%%%%%%%%%%%%%%%%%%%%%%%%%%%%%%%%%%%%%%%%%%%%%%%%%%%%%%%%%%%%%%
% Required packages:
% \usepackage{booktabs}
% \usepackage{graphicx}

\section{Experiments}
\label{sec:exps}

\subsection{Experimental Setup}

\paragraph{\textbf{Benchmark suite}}We illustrate our protocol on the LIBERO benchmark containing data suites - LIBERO-Goal, LIBERO-Spatial, LIBERO-Object, and LIBERO-Long. Each data suite contains ten robot tasks presented sequentially. We use \textbf{LIBERO-Goal} for the main evaluations because its tasks span diverse goals involving changes in actions, objects, targets, and spatial relations, providing broad coverage for evaluating semantic invariance and instruction-dependent behavioral changes. At each continual learning task $t$, the policy is trained on task $t$ and evaluated on all previously observed tasks $j \leq t$.

\paragraph{\textbf{Compared methods}}We evaluate eight representative continual imitation learning approaches. Sequential Fine-Tuning (SeqFT), Experience Replay (ER), Elastic Weight Consolidation (EWC), PackNet, DMPEL, IsCiL, L2M, and TAIL. SeqFT provides a lower-bound baseline without a specific mechanism to mitigate catastrophic forgetting. ER has a fixed size buffer to store representative demonstrations from previous tasks through reservoir update and random replay~\cite{ER}, EWC
regularizes parameters according to their importance, and
PackNet assigns task-specific parameter subsets through iterative pruning. L2M, IsCiL, TAIL, and DMPEL employ parameter-efficient or modular adaptation mechanisms. We preserve each method's native routing mechanism and report any use of task identity.

% ------------------------------------------------------------
% TABLE 1: SEMANTIC INVARIANCE
% ------------------------------------------------------------
\begin{table*}[!t]
\centering
\caption{Semantic invariance on \textbf{LIBERO-Goal}. Metrics include Original AUC ($\uparrow$), Paraphrase AUC ($\uparrow$), signed paraphrase gap $\Delta_{\mathrm{AUC}}^{\mathrm{para}}$, Original Final SR ($\uparrow$), and Original NBT ($\downarrow$). We establish the standard deviation across three seeds.}
\label{tab:goal_semantic_invariance}
\begin{tabular}{lccccc}
\toprule
Method
& Original AUC $\uparrow$
& Paraphrase AUC $\uparrow$
& $\Delta_{\mathrm{AUC}}^{\mathrm{para}}$
& Original Final SR $\uparrow$
& Original NBT $\downarrow$ \\
\midrule
L2M     & 7.06 $\pm$ 0.02 & 5.44 $\pm$ 0.02 & 1.62 $\pm$ 0.04 & 5.01 $\pm$ 0.01 & 8.44 $\pm$ 0.05 \\
SeqFT   & 24.63 $\pm$ 0.02 & 19.18 $\pm$ 0.03 & 5.45 $\pm$ 0.05 & 8.00 $\pm$ 0.02 & 76.49 $\pm$ 1.04 \\
ER      & $58.54 \pm 0.18$ & $51.82 \pm 0.27$ & $6.72 \pm 0.11$ & $59.23 \pm 0.24$ & $17.14 \pm 0.09$ \\
EWC     & $32.77 \pm 0.13$ & $27.46 \pm 0.21$ & $5.31 \pm 0.07$ & $27.41 \pm 0.29$ & $46.18 \pm 0.16$ \\
PackNet & $53.34 \pm 0.25$ & $46.93 \pm 0.14$ & $6.41 \pm 0.19$ & $53.27 \pm 0.08$ & $15.36 \pm 0.31$ \\
DMPEL   & $78.33 \pm 0.12$ & $70.28 \pm 0.18$ & $8.05 \pm 0.22$ & $81.12 \pm 0.17$ & $0.00 \pm 0.00$ \\
IsCiL   & $60.72 \pm 0.28$ & $54.67 \pm 0.12$ & $6.05 \pm 0.20$ & $53.44 \pm 0.26$ & $5.17 \pm 0.08$ \\
TAIL    & $67.39 \pm 0.15$ & $61.18 \pm 0.30$ & $6.21 \pm 0.09$ & $64.31 \pm 0.22$ & $22.26 \pm 0.13$ \\
\bottomrule
\end{tabular}
\vspace{-10pt}
\end{table*}

\paragraph{\textbf{Policy and Optimization}} All methods are initialized from the same pretrained multitask policy
and use a common behavior-cloning architecture. The policy contains CLIP-based visual and language encoders, a six-layer transformer with eight attention heads, and a Gaussian-mixture action head. We train each task for 10 epochs using AdamW with an initial learning rate of $10^{-4}$, weight decay of $0.1$, and a cosine scheduler. The training and evaluation batch sizes are 32 and 8, respectively. Training uses image color, translation, and affine augmentation. Method specific hyperparameters follow their respective public implementations. ER uses a memory capacity of 1,000 demonstrations, EWC uses $\lambda=50{,}000$ and $\gamma=0.9$, and PackNet prunes
$75\%$ of the currently available parameters before running post pruning training.

All experiments are performed using PyTorch on a single NVIDIA H100 GPU. MuJoCo simulation is rendered using EGL, and the same rendering backend, task order, pretrained initialization, and optimization configuration are used across methods. We use three random seeds: 101, 102, and 103.

\noindent\textbf{Evaluation protocol.}
Policies are trained exclusively using the original LIBERO benchmark instructions. We save a checkpoint after every task and evaluate the frozen checkpoints separately, ensuring that diagnostic evaluation does not modify the policy or continual-learning state.

We use the full instruction profile, which evaluates every retained paraphrase, minimal contrast, and statically validated collision in the manifest. Each instruction variant is evaluated using 20 simulator initial states. Paraphrases are evaluated at every continual-learning task to measure temporal semantic invariance. Minimal contrasts and collisions are evaluated after the final task. Original and perturbed conditions therefore use the same rollout budget of 20 initial states per evaluated instruction.
%%%%%%%%%%%%%%%%%%%%%%%%%%%%%%%%%%%%%%%%%%%%%%%%%%%%%%%%%%%%%%%%%%%%%%%%
%%%%%%%%%% RESULTS   %%%%%%%%%%%%%%%%%%%%%%%%%%%%%%%%%%%%%%%%%%%%%%%%%%
%%%%%%%%%%%%%%%%%%%%%%%%%%%%%%%%%%%%%%%%%%%%%%%%%%%%%%%%%%%%%%%%%%%%%%%%
\section{Results and Discussion}
\label{sec:results}

\subsection{Semantic Invariance Across Learning Stages}
\label{sec:semantic_results}

Table~\ref{tab:goal_semantic_invariance} compares performance under original and meaning-preserving paraphrased instructions throughout the continual learning tasks. Original performance measures how well each method learns and retains the manipulation tasks, while paraphrase performance tests whether this behavior remains stable under changes in sentence structure. Methods with stronger continual-learning performance generally perform better under paraphrasing. DMPEL achieves the highest Original and Paraphrase AUC, at 78.51 and 71.24, followed by TAIL and IsCiL in Paraphrase AUC, then ER and PackNet. This suggests that better task retention also preserves representations that remain useful under instruction variation.

All methods, however, degrade under paraphrasing. The AUC drop ranges from 1.61 points for L2M to 7.27 points for DMPEL. L2M's small gap does not indicate robustness because its overall performance is low, whereas DMPEL has the largest gap but still achieves the best Paraphrase AUC. The gap must therefore be interpreted together with absolute performance.

The results further show that continual-learning performance and language robustness are distinct. DMPEL achieves the strongest Original Final SR and zero NBT, yet still exhibits a nonzero paraphrase gap. ER, PackNet, IsCiL, and TAIL also retain more task performance than SeqFT while losing roughly 6--7 AUC points under paraphrasing. Thus, reducing forgetting does not necessarily make the policy invariant to semantically equivalent instructions.

Although all methods share the same language-conditioned policy backbone, the continual-learning strategy changes how this backbone evolves. Replay-based methods retain broader language-task associations and achieve higher absolute Paraphrase AUC, while sequential fine-tuning is more susceptible to representation drift. Parameter-isolation and regularization methods preserve prior behavior but do not explicitly enforce similar representations for semantically equivalent instructions. We therefore treat high Paraphrase AUC together with a small signed gap, as stronger evidence of semantic invariance than the gap alone.

% ------------------------------------------------------------
% TABLE 2: EXECUTABLE MINIMAL CONTRAST — OVERALL RESULTS
% ------------------------------------------------------------
\begin{table}[t]
\centering
\caption{Final-stage executable minimal-contrast results on
\textbf{LIBERO-Goal}. Higher Target SR and GSA are better, while lower
Original Persistence is better.}
\footnotesize
\label{tab:goal_minimal_contrast}
\begin{tabular}{lccc}
\toprule
Method
& Target SR $\uparrow$
& Average GSA $\uparrow$
& Original Persistence $\downarrow$ \\
\midrule
L2M     & 4.42  & 2.81  & 4.61  \\
SeqFT   & 7.35  & 4.41  & 8.82  \\
ER      & 19.42 & 11.73 & 37.84 \\
EWC     & 10.68 & 6.57  & 20.81 \\
PackNet & 15.76 & 8.24  & 42.63 \\
DMPEL   & \textbf{30.91} & \textbf{21.68} & 44.72 \\
IsCiL   & 21.37 & 13.92 & 32.81 \\
TAIL    & 24.86 & 15.79 & 37.62 \\
\bottomrule
\end{tabular}
\vspace{-10pt}
\end{table}

\subsection{Response to Minimal Semantic Changes}
\label{sec:minimal_results}

Table~\ref{tab:goal_minimal_contrast} shows that strong benchmark performance does not necessarily imply reliable goal switching. DMPEL achieves the highest Target SR (30.91) and GSA (21.68), followed by TAIL, IsCiL, and ER, while L2M and SeqFT have GSA below 4\%. Target SR is consistently higher than GSA because reaching the modified target does not always mean abandoning the original goal. DMPEL, for example, still has an Original Persistence of 44.72. This indicates that even strong methods often remain influenced by the original task association. 

\begin{figure}[t]
\centering
%\vspace{-10pt}
\includegraphics[width=0.95\columnwidth]{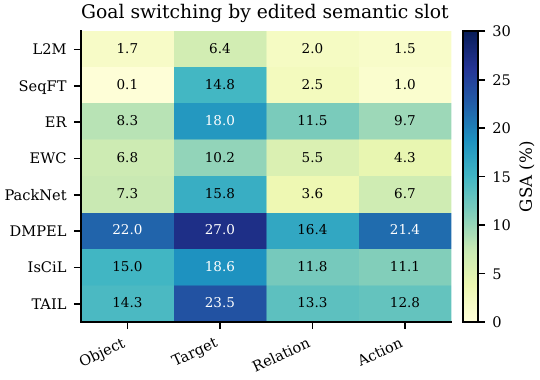}
\caption{Goal-Switch Accuracy (GSA) by edited semantic slot on \textbf{LIBERO-Goal}. Higher values indicate better adaptation to the modified instruction. No executable attribute contrasts were generated.}
\vspace{-15pt}
\label{fig:gsa_by_slot}
\end{figure}

Additionally, Figure~\ref{fig:gsa_by_slot} shows that sensitivity also depends on the edited semantic slot. Target edits achieve the highest average GSA ($\sim$16.78\%), while object, relation, and action edits remain around 8 to 9\%. DMPEL responds more consistently across slots, whereas SeqFT responds mainly to target changes. No attribute results are reported because the benchmark contains no retained attribute minimal contrasts.
% ------------------------------------------------------------
% FIGURE: GSA BY EDITED SEMANTIC SLOT
% ------------------------------------------------------------

% ------------------------------------------------------------
% FIGURE: COLLISION BEHAVIOR AND ACTION-LEVEL SENSITIVITY
% ------------------------------------------------------------
\begin{figure*}[!b]
\centering
%\vspace{-10pt}
\includegraphics[width=0.8\textwidth]{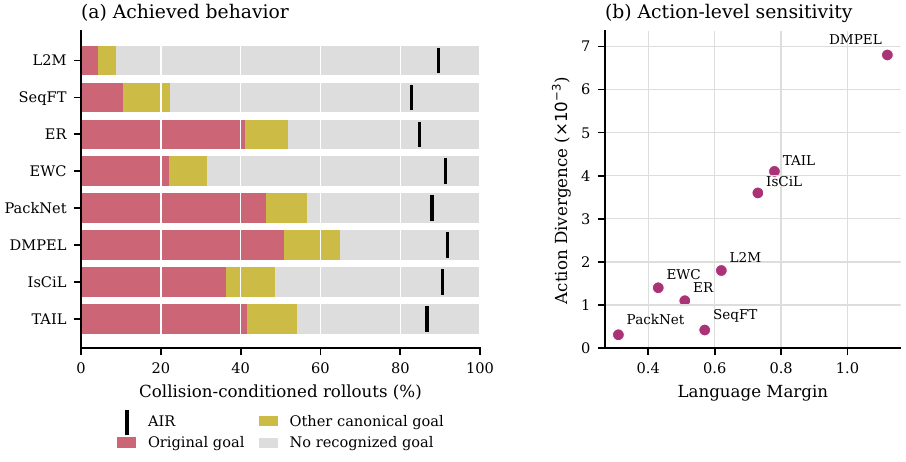}
%\vspace{-1pt}
\caption{Final-stage behavior under scene-incompatible instructions on\textbf{LIBERO-Goal} Left: the proportions of rollouts achieving the original goal, another audited canonical goal, or no recognized goal; vertical markers indicate the Action-Initiation Rate (AIR). Right: Language Margin versus Action Divergence, characterizing action-level sensitivity to the instruction perturbation.}
%\vspace{-10pt}
\label{fig:goal_collision_behavior}
\end{figure*}

Table~\ref{tab:goal_minimal_contrast} evaluates whether the final policy responds to a controlled change in one semantic component. Target success measures whether the modified goal is reached, whereas GSA additionally requires that the original goal is not achieved. The latter distinction is important because a rollout may satisfy both goal predicates or may continue executing the original task despite the modified instruction.

The semantic-slot breakdown characterizes performance under different types of instruction changes. Performance differences between \textbf{object}, \textbf{target}, \textbf{relation}, \textbf{action}, and \textbf{attribute} edits reveal whether a method grounds the complete instruction or relies primarily on a subset of its semantic content.

\subsection{Behavior Under Incompatible Instructions}
\label{sec:collision_results}

Figure~\ref{fig:goal_collision_behavior} examines final-stage behavior under statically validated scene-incompatible instructions. Original Final SR provides the reference, since low Original-Goal Persistence (OGP) is only meaningful when a method can reliably complete the original task. At the output, incompatible instructions rarely cause policies to stop acting, with AIR remaining between 83 to 92\% across methods. Stronger methods are also more likely to fall back to a previously learned goal. DMPEL has the highest CGAR at 64.90\% and reaches the original goal in 50.80\% of rollouts, while L2M's lower OGP and CGAR are partly explained by weaker original-task performance. At the action level, DMPEL also shows the largest Language Margin and Action Divergence, followed by TAIL and IsCiL. Overall, these results show that action-level sensitivity does not always produce a behavioral switch, motivating both outcome and action-level diagnostics. These metrics are not treated as collision success rates because the incompatible instruction has no valid target goal.

% ------------------------------------------------------------
% INSTRUCTION COVERAGE
% ------------------------------------------------------------
\subsection{Instruction-Manifest Coverage}
\label{sec:coverage_results}
\begin{figure}[h!]
    \centering
    \vspace{-15pt}
    \includegraphics[width=0.75\linewidth]{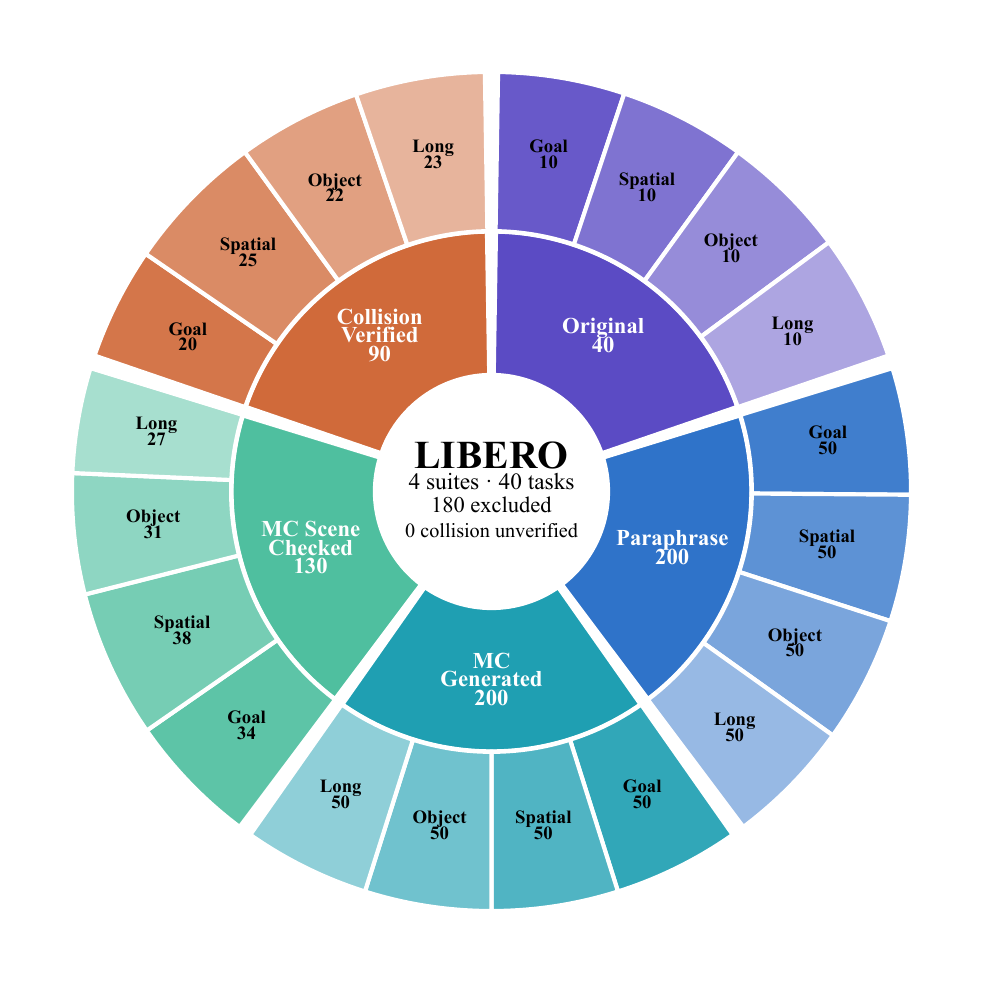}
    \vspace{-10pt}
    \caption{Distribution of the instruction perturbations used to evaluate language grounding across the four LIBERO suites. The figure summarizes the original tasks, paraphrases, generated minimal contrasts, scene-validated minimal contrasts, and scene-incompatible collision instructions, together with their semantic categories.}
    \label{fig:data_distribution}
    \vspace{-15pt}
\end{figure}

Figure~\ref{fig:data_distribution} summarizes the instruction variants retained by the construction and validation pipeline across the four LIBERO suites. All 40 original instructions and 200 paraphrases are retained, while 130 of 200 minimal contrasts and 90 of 200 scene-incompatible collisions pass the required validation checks. Overall, 460 variants are retained and 180 generated candidates are excluded.

Coverage varies across conditions because minimal contrasts and collisions modify semantic content and may introduce unsupported objects, relations, or goal objective. Such variants are excluded rather than assigned an unsupported evaluation target. This preserves valid one-slot interventions while naturally producing different eligible subsets across tasks and semantic slots.

%%%%%%%%%%%%%%%%%%%%%%%%%%%%%%%%%%%%%%%%%%%%%%%%%%%%%%%%%%%%%%%%%%%%%%%%
%%%%%%%%%% ABLATIONS  %%%%%%%%%%%%%%%%%%%%%%%%%%%%%%%%%%%%%%%%%%%%%%%%%%
%%%%%%%%%%%%%%%%%%%%%%%%%%%%%%%%%%%%%%%%%%%%%%%%%%%%%%%%%%%%%%%%%%%%%%%%
\section{Additional Diagnostic Analysis}

\subsection{Effect of Temporal Evaluation}
\label{sec:temporal_analysis}

\begin{figure}[t]
\centering
\includegraphics[width=0.85\columnwidth]{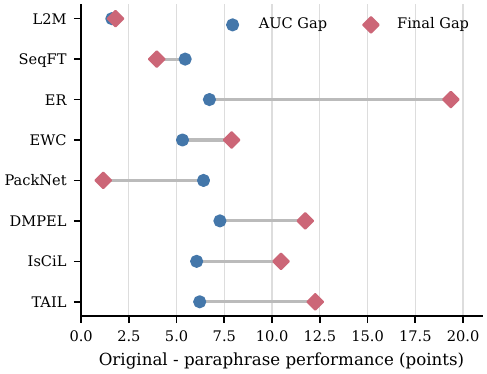}
\caption{Paraphrase sensitivity at the final stage and throughout
continual learning. Final Gap measures the difference between original and paraphrase success at the final stage, whereas AUC Gap measures the corresponding difference across all stages. Lower values indicate greater semantic invariance.}
\vspace{-15pt}
\label{fig:temporal_analysis}
\end{figure}

We examine whether evaluating only the final checkpoint provides the same conclusions as evaluation throughout the continual-learning sequence. Figure~\ref{fig:temporal_analysis} shows that final-only and temporal evaluation can tell different stories. PackNet has the smallest Final Gap (1.16) but a larger AUC Gap (6.41), suggesting that it recovers from earlier sensitivity. In contrast, ER has a large Final Gap (19.36) but a smaller AUC Gap (6.72), with similar trends for DMPEL, IsCiL, and TAIL. This shows that final-only evaluation can miss earlier instability, while temporal averages can hide poor final-stage performance.

Figure~\ref{fig:temporal_analysis} shows that final-only and temporal evaluation can produce substantially different conclusions. PackNet has the smallest Final Gap, at 1.16 points, despite an AUC Gap of 6.41 points. Its final policy responds similarly to original and paraphrased instructions, but this invariance is not maintained throughout the learning sequence. L2M also has similar final and temporal gaps, although both are measured at a low absolute level of
task performance.

ER exhibits the largest discrepancy in the opposite direction: its Final Gap is 19.36 points, compared with an AUC Gap of 6.72 points. DMPEL, IsCiL, and TAIL exhibit the same pattern, with final-stage gaps substantially exceeding their temporal gaps. For these methods, stage-averaged evaluation understates the paraphrase sensitivity of the final policy. These results demonstrate that final-only evaluation can conceal earlier instability, while temporal averages can conceal pronounced degradation at the final checkpoint.

% ------------------------------------------------------------
% FIGURE: TEMPORAL VERSUS FINAL PARAPHRASE GAP
% ------------------------------------------------------------

\subsection{Complementarity of Collision Diagnostics}
\label{sec:diagnostic_complementarity}

We examine whether action-level sensitivity can substitute for outcome-level evaluation by measuring Spearman's rank correlation~\cite{spearman1904proof} across eligible method-task pairs. Specifically, we compare Action Divergence (AD) and Language Margin (LM) with Original-Goal Persistence (OGP), and Action-Initiation Rate (AIR) with Canonical-Goal Activation Rate (CGAR).

Table~\ref{tab:diagnostic_complementarity} shows that AIR and CGAR are strongly associated ($\rho=0.80$), indicating that policies that initiate behavior more often are also more likely to activate a canonical goal. However, this does not imply correct instruction following because collision instructions have no valid scene goal.

\begin{table}[h!]
\centering
\footnotesize
\caption{Association between action-level and outcome-level collision
diagnostics across method-task pairs. Weak associations indicate
that the measurements capture complementary policy behaviors.}
\label{tab:diagnostic_complementarity}
\begin{tabular}{lc}
\toprule
Diagnostic Pair & Spearman $\rho$ \\
\midrule
AD vs.\ OGP   & 0.26 \\
AIR vs.\ CGAR & 0.80 \\
LM vs.\ OGP   & 0.21 \\
\bottomrule
\end{tabular}

\end{table}

In contrast, AD and LM are only weakly associated with OGP ($\rho=0.26$ and $0.21$). Thus, changes in the action distribution do not reliably predict changes in achieved behavior. These results motivate reporting both action- and outcome-level diagnostics, as they capture complementary aspects of policy response.
%%%%%%%%%%%%%%%%%%%%%%%%%%%%%%%%%%%%%%%%%%%%%%%%%%%%%%%%%%%%%%%%%%%%%%%%
%%%%%%%%%% CONCLUSION %%%%%%%%%%%%%%%%%%%%%%%%%%%%%%%%%%%%%%%%%%%%%%%%%%
%%%%%%%%%%%%%%%%%%%%%%%%%%%%%%%%%%%%%%%%%%%%%%%%%%%%%%%%%%%%%%%%%%%%%%%%
\section{Limitations and Conclusion}

We introduced a temporal diagnostic protocol for testing whether continual robotic policies remain grounded in language as they acquire new tasks. Across LIBERO, conventional task retention did not guarantee semantic invariance, reliable goal switching, or appropriate behavior under scene-incompatible instructions. The results show that paraphrase, minimal-contrast, and collision diagnostics expose complementary limitations that task success alone cannot identify.

Our study uses a common behavior-cloning architecture and does not evaluate large-scale VLAs, whose broader pretraining may produce different behavior. Minimal-contrast and collision diagnostics are evaluated only at the final checkpoint, so they do not isolate changes caused by continual learning from pre-existing limitations. Failures on modified goals may also reflect limited task-execution capability rather than language interpretation alone. These diagnostics therefore characterize behavior without establishing a causal mechanism for grounding failure.

Despite these limitations, the proposed protocol provides a more detailed view of language grounding than task success alone. We hope this benchmark supports more rigorous evaluation of language grounding in continual robot learning.

\section*{Acknowledgments}
We used OpenAI's ChatGPT for generating the robotic arm images in Figure~\ref{fig:benchmark}, and to assist with writing. No scientific or technical content was generated using AI tools.
% ------------------------------------------------------------
% References
% ------------------------------------------------------------
\bibliographystyle{IEEEtran}
\bibliography{IEEEabrv, references}

% Appendix
% \input{sections/8_appendix}
% ============================================================
\end{document}